\pdfoutput=1
\documentclass[conference]{IEEEtran}
\IEEEoverridecommandlockouts
\usepackage{cite}
\usepackage{amsmath,amssymb,amsfonts}
\usepackage{graphicx}
\DeclareGraphicsExtensions{.pdf,.png,.jpg}
\usepackage{textcomp}
\usepackage{xcolor}
\def\BibTeX{{\rm B\kern-.05em{\sc i\kern-.025em b}\kern-.08em
    T\kern-.1667em\lower.7ex\hbox{E}\kern-.125emX}}

\usepackage{fancyhdr}
\usepackage{enumitem}
\usepackage{booktabs}
\usepackage{url}
\usepackage[colorlinks=true,
            linkcolor=black,   % section links
            citecolor=blue,    % citation links
            urlcolor=blue]{hyperref}  % <-- load LAST

\usepackage{tablefootnote}
\usepackage{listings}
\usepackage{algorithm}
\usepackage{algpseudocode}
\usepackage{subcaption}
\begin{document}

\title{Ensembling Large Language Models to Characterize Affective Dynamics in Student–AI Tutor Dialogues\\

}

\author{\IEEEauthorblockN{Chenyu Zhang}
\IEEEauthorblockA{\textit{Harvard Graduate School of Education} \\
\textit{Harvard University} \\
Cambridge, MA, USA \\
chenyu\_zhang@alumni.harvard.edu}
\and
\IEEEauthorblockN{Sharifa Alghowinem}
\IEEEauthorblockA{\textit{Personal Robots Group, Media Lab} \\
\textit{Massachusetts Institute of Technology} \\
Cambridge, MA, USA \\
sharifah@media.mit.edu}
\and
\IEEEauthorblockN{Cynthia Breazeal}
\IEEEauthorblockA{\textit{Personal Robots Group, Media Lab} \\
\textit{Massachusetts Institute of Technology} \\
Cambridge, MA, USA \\
cynthiab@media.mit.edu}
}

\maketitle
\thispagestyle{fancy}

\begin{abstract}
While recent studies have examined the leaning impact of large language model (LLM) in educational contexts, the affective dynamics of LLM-mediated tutoring remain insufficiently understood. This work introduces the first ensemble-LLM framework for large-scale affect sensing in tutoring dialogues, advancing the conversation on responsible pathways for integrating generative AI into education by attending to learners’ evolving affective states. To achieve this, we analyzed two semesters' worth of 16,986 conversational turns exchanged between PyTutor, an LLM-powered AI tutor, and 261 undergraduate learners across three U.S. institutions. 
To investigate learners' emotional experiences, we generate zero-shot affect annotations from three frontier LLMs (Gemini, GPT-4o, Claude), including scalar ratings of valence, arousal, and learning–helpfulness, along with free-text emotion labels.
These estimates are fused through rank-weighted intra-model pooling and plurality consensus across models to produce robust emotion profiles.
% Results reveal that distributions center on mildly positive valence (\(\tilde v\!\approx\!5\)), moderate arousal (\(\tilde a\!=\!5\text{--}6\)), and above-neutral learning (\(\tilde \ell\!\approx\!6\)), yet affective friction persists (confusion \(22.15\%\); frustration \(8.62\%\)). Affect shows short-horizon persistence: positive states are most stable (mean dwell \(\approx 2.3\) turns), negative less so (\(2.0\)), and neutral brief (\(1.4\)). Over half of negative turns repair immediately, with direct rebounds to positive (\(0.29\)) more common than routes via neutral (\(0.22\)); neutral acts as a gateway that more often tips upward. Still, positive momentum is fragile, with a \(43\%\) chance of reverting to non-positive next turn.
Our analysis shows that during interaction with the AI tutor, students typically report mildly positive affect and moderate arousal. Yet learning is not uniformly smooth: confusion and curiosity are frequent companions to problem solving, and frustration, while less common, still surfaces in ways that can derail progress. Emotional states are short-lived—positive moments last slightly longer than neutral or negative ones, but they are fragile and easily disrupted. Encouragingly, negative emotions often resolve quickly, sometimes rebounding directly into positive states. Neutral moments frequently act as turning points, more often steering students upward than downward, suggesting opportunities for tutors to intervene at precisely these junctures.
\end{abstract}


\section{Introduction}
\label{sec:intro}
% -------------------------------------------------------------
% Hook & motivation
Large language model (LLM)–based tutors promise on-demand hints,
dialogue-level adaptation, and scalable personalization.  Yet recent
classroom studies report mixed or even negative effects on learning
gains \cite{Bastani2024,Kosmyna}.  Most evaluations still judge success
solely by post-test scores or neurophysiological proxies such as
EEG-derived cognitive load, leaving the felt emotional experience
of learners virtually unmeasured, even though emotion steers attention
\cite{Vuilleumier2005}, memory \cite{Phelps2004,Um2012}, and reasoning
\cite{Jung2014}.  In practice, students often quit when frustration
mounts; an AI tutor’s real success may lie in navigating those
emotions through adaptive tone, timely feedback, and encouragement.

% -------------------------------------------------------------
% Prior work & positioning
Affective science offers two complementary lenses.  One maps emotions to
a continuous valence–arousal–dominance (VAD) space \cite{Mehrabian1980};
the other lists biologically primed “basic” categories
\cite{Ekman1999}.  Hybrid accounts bridge the views \cite{Ellis2012}.
Extending them to education, Kort’s learning–unlearning spiral adds a
progress axis that rates how constructive an emotion is for mastery
\cite{kort-2001}.  
Building on models of affective dynamics that link learner emotions to comprehension progress \cite{DMello2014}, we operationalize a third axis, learning–helpfulness, using a 1–9 Likert scale to index how much each conversational turn advances a student’s understanding.

Yet the corpora driving most emotion models were tagged by only a few
annotators (typically 3), neglect cultural variance, and force single‐label
decisions—choices at odds with mixed-emotion findings
\cite{Berrios2015} and liable to propagate bias downstream. LLMs, trained on vast and diverse
corpora, can dilute some of that bias, yet introduce their own
idiosyncrasies \cite{llm-biases}. Despite such advances,
large-scale evidence on the \emph{affective} impact of LLM tutors
remains scarce; most studies still track purely cognitive metrics. We therefore adopt a
\emph{three-model ensemble}: independent inferences from Gemini, GPT-4o,
and Claude are fused via hierarchical consensus, reducing both
annotator- and model-specific errors.  
% -------------------------------------------------------------
% Research questions
We ask:

\begin{enumerate}[label=\textbf{RQ\arabic*}, leftmargin=*, align=left]
  \item Which affective states, both scalar and categorical, dominate
        student–AI tutor interactions?
  \item How do these states evolve over time, especially with respect
        to valence transitions?
\end{enumerate}
% -------------------------------------------------------------
% Contribution & novelty
We answer them by analyzing 16,986 conversational turns collected
across two semesters from \emph{PyTutor}~\cite{pytutor}, a GPT-4o–based Socratic
AI tutor, used by 261 undergraduates at the Massachusetts Institute of Technology (MIT), Georgia State University (GSU), and Quinsigamond Community College (QCC).
Three frontier models—Gemini \cite{b1}, GPT-4o \cite{b2},
and Claude \cite{b3}—provide zero-shot estimates of valence, arousal,
and learning-helpfulness (VAL) plus free-text emotion labels.  
We fuse those signals through rank-weighted intra-model pooling and cross-model plurality consensus to obtain robust, turn-level affect annotations.

Overall, this work delivers the first large-scale portrait of
affective dynamics in LLM-mediated tutoring, bridging a critical gap
between cognitive and emotional evaluation of AI education tools.

\section{Method}
\label{sec:method}

Building on \textbf{RQ1} and \textbf{RQ2}, this section details the
dataset, preprocessing, and ensemble-annotation pipeline that yield
turn-level affect labels.  We then outline the temporal analyses used
to trace valence transitions and answer our research questions.

\subsection{Dataset}
The dataset for this study comes from \textit{PyTutor}, a GPT-4o–based Socratic AI tutor deployed during the Fall 2024 and Spring 2025 semesters at MIT, GSU, and QCC.
PyTutor provides on-demand hints, withholds complete solutions, and prompts learners to debug their own code via metacognitive questions. The deployment involved \textbf{261} undergraduates enrolled in introductory Python or computing courses, all working in English.

Across two 15-week semesters these learners produced \textbf{16,986} human–AI conversational turns. A “session’’ is defined as a contiguous block of activity separated by 60 min of inactivity. Table \ref{tab:data-stats} summarizes key statistics for participants, sessions, turns, and tokens.

\begin{table}[!t]
\centering
\caption{PyTutor corpus summary. “Overall” values describe the entire dataset;
“Per–participant” values are the median \emph{(inter-quartile range)} across
the 261 learners.}
\label{tab:data-stats}
\begin{tabular}{lcc}
\toprule
\textbf{Metric} & \textbf{Overall} & \textbf{Per–participant} \\
\midrule
Participants                              & 261                     & -- \\
Sessions                                  & 1,486                   & 1.0\,(1.0–4.0) \\
Conversation span (days)                  & 274                     & 0.0\,(0.0–29.0) \\
Days active                               & 194                     & 1\,(1–4) \\
Turns                                     & 16,986                  & 16\,(6–46) \\
Turns per session                         & 6.0\,(2.0–14.0)         & 8.7\,(4.0–14.0) \\
Tokens                                    & 2,113,485               & 1,628\,(529–5\,173) \\
Tokens per turn                           & 57\,(9–221)             & 99.7\,(69.1–133.2) \\
Code snippets                             & 19,782                  & 9\,(0–42) \\
Code snippets per turn                    & 1.165                   & 0.462 (0.000–1.299) \\
Session duration (min)                    & 36,636.6                & 12.4\,(2.0–29.5) \\
First activity                            & 2024-09-24              & -- \\
Last activity                             & 2025-06-26              & -- \\
\bottomrule
\end{tabular}
\end{table}

% -------------------------------------------------------------------------
\subsection{Automatic Emotion Annotation \& Ensemble Fusion}
\label{sec:annotation-fusion}

Manual affect annotation is notoriously burdensome: a single
crowdsourced label set can cost tens of hours, typically relies on only
a few annotators, and inherits their cultural and taxonomic biases while
omitting learning-specific constructs altogether.  These constraints
make it impractical to obtain turn-level ground truth at
semester scale.  We therefore opt for \emph{automatic} annotation via an
ensemble of frontier LLMs, leveraging their broad pre-training to
minimize single-annotator bias while still acknowledging model
idiosyncrasies. We generate turn-level affect annotations through a two-step pipeline that
combines \textbf{(i) zero-shot emotion inference} from multiple LLMs with
\textbf{(ii) hierarchical ensemble fusion}.

\paragraph*{Zero-shot annotation.}
Every student–tutor utterance is processed by three frontier
models: Gemini 2.0 Flash (Google), GPT-4o mini (OpenAI), and Claude 3.5 Sonnet
(Anthropic), under an identical system prompt.  
Each model returns \emph{two} outputs:  
(i) a ranked list of up to $K=5$ discrete emotion labels, and  
(ii) three 1–9 Likert ratings for \emph{valence} ($v$), \emph{arousal} ($a$),
and \emph{learning-helpfulness} ($\ell$).  
Apart from the placeholder \texttt{``neutral''}, no candidate labels are
hard-coded, enabling the models to surface domain-specific states such as
\emph{puzzlement} or \emph{anticipation}.

\paragraph*{Hierarchical ensemble fusion.}
Let $m\!\in\!\{1,2,3\}$ index models and
$r\!\in\!\{1,\dots ,K_m\}$ the rank within a model’s output.
Fusion proceeds in three stages:

\begin{description}[leftmargin=1.7em]
  \item[Stage 1: Intra-model pooling.]
        Rank $r$ receives a linearly decaying weight
        $w_r\propto K_m-r+1$ (normalized so $\sum_r w_r=1$).
        Weighted means yield per-model scores
        \[
          \hspace{-1.5em}
          \hat v_{m,t}= \sum_{r} w_r\,v_{m,t}^{(r)},\
          \hat a_{m,t}= \sum_{r} w_r\,a_{m,t}^{(r)},\
          \hat\ell_{m,t}= \sum_{r} w_r\,\ell_{m,t}^{(r)}.
        \]

  \item[Stage 2: Inter-model aggregation.]
        Scalars are averaged across the $M=3$ models:
        \[
          \bar v_t=\tfrac1M\sum_{m}\hat v_{m,t},\
          \bar a_t=\tfrac1M\sum_{m}\hat a_{m,t},\
          \bar\ell_t=\tfrac1M\sum_{m}\hat\ell_{m,t}.
        \]

  \item[Stage 3: Label consensus.]
        For discrete labels, let $f_t(e)$ be the number of models that emit
        label~$e$.  
        The primary ensemble label is the plurality winner
        $e_t^{\star}=\arg\max_e f_t(e)$, with ties broken by higher
        $\bar v_t$ and, if still tied, lexicographic order.
\end{description}

Each conversational turn~$t$ thus receives a fused VAL triplet
$\bigl(\bar v_t,\bar a_t,\bar\ell_t\bigr)$ and one consensus emotion
label, which underpin the analyses in
Sections \ref{sec:results}–\ref{sec:discussion}.

% \smallskip
% \noindent\textbf{Limitations.}
% LLM–derived VAL scores and emotion labels are \emph{proxies}; without a
% human-annotated subset the ensemble may simply average correlated noise
% rather than converge to ground truth.  Additionally, our arithmetic means assume the
% $1\!-\!9$ Likert scale is interval, which can misrepresent central
% tendency if it is merely ordinal.  Finally, the linearly decaying
% weights $w_r\!\propto\!K_m-r+1$ and equal model weights are heuristic;
% alternative schemes (e.g., learned per-model reliabilities or ordinal
% regression) may yield different conclusions.

% ============================================================
\subsection{Temporal Analyses}
\label{sec:temporal}
% ============================================================
\paragraph*{Markov transition dynamics.}
To analyze inter-state flow, we model discrete affect transitions as a \emph{first-order Markov chain} with state space $\{\textit{positive},\textit{neutral},\textit{negative}\}$, based on valence tertiles computed globally across all student turns. Transitions are computed within sessions (separated by $\geq$60 minutes of inactivity), with the first turn of each session excluded to avoid initialization bias. We estimate the $3\times3$ transition matrix $\mathbf{P}$ via frequency counts with Laplace smoothing ($\beta=1$), and compute expected dwell time in each state as $\text{dwell}(s)=1/(1-P_{ss})$.

%%%%%%%%%%%%%%%%%%%%%%%%%%%%%%%%%%%%%%%%%%%%%%%%%%%%%%%%%%%%%%%%
% ======================  R E S U L T S  =======================
%%%%%%%%%%%%%%%%%%%%%%%%%%%%%%%%%%%%%%%%%%%%%%%%%%%%%%%%%%%%%%%%
\section{Results}
\label{sec:results}
\noindent
\noindent
With the overarching goal of characterizing human affective
dynamics, all main-text analyses below focus on \textbf{student turns
only}.
% --------------------------------------------------------------
\subsection{Descriptive Affective Landscape}
\label{sec:landscape}

\begin{figure*}[!t]
  \centering
  \begin{subfigure}[t]{0.49\linewidth}
    \includegraphics[width=\linewidth]{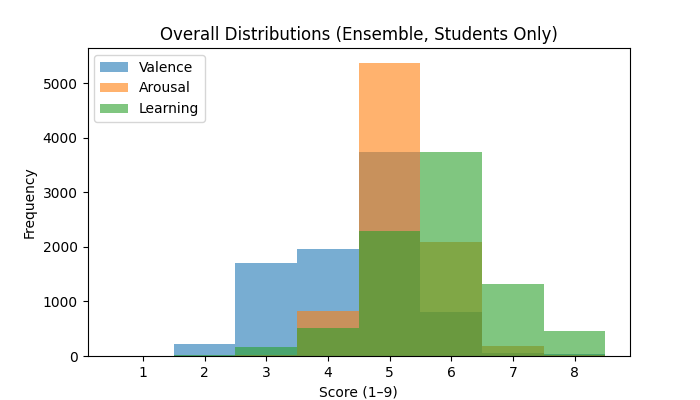}
    \caption{Stacked histograms of ensemble valence ($v$), arousal ($a$),
             and learning ($\ell$) scores for all student turns
             ($n=16,986$).}
    \label{fig:overall_dists}
  \end{subfigure}\hfill
  \begin{subfigure}[t]{0.49\linewidth}
    \includegraphics[width=\linewidth]{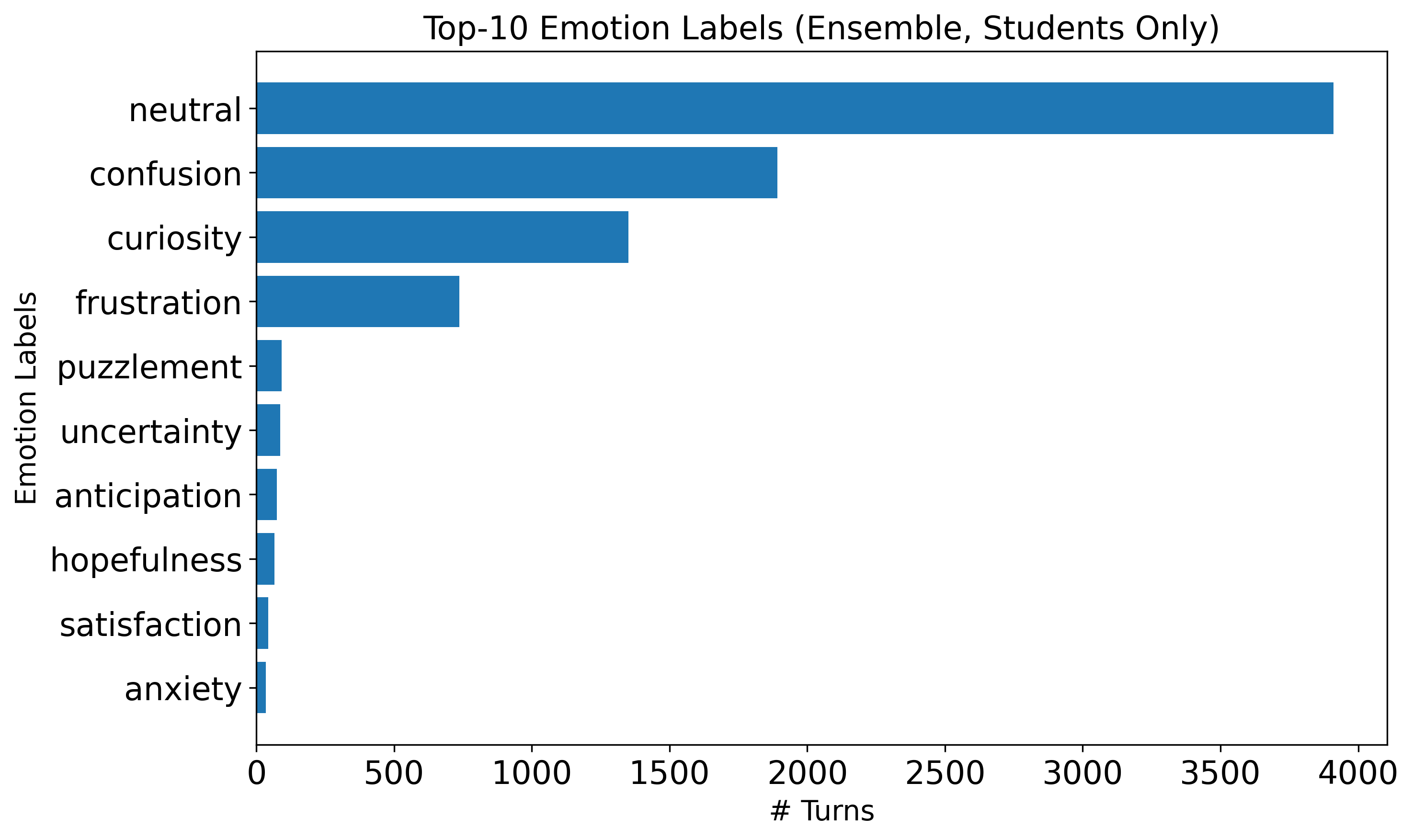}
    \caption{Frequency of the ten most common ensemble emotion labels.}
    \label{fig:top10_emotions}
  \end{subfigure}
  \caption{Affective profile of \emph{PyTutor} student turns.}
  \label{fig:affective_landscape}
\end{figure*}

\paragraph*{Scalar VAL distributions (students only).}
Figure~\ref{fig:overall_dists} overlays the ensemble‐fused
\emph{valence} ($v$), \emph{arousal} ($a$), and
\emph{learning-helpfulness} ($\ell$) scores for all 16,986 student
turns.  
The distributions are tightly centered:  
$\operatorname{median}v=5$ (IQR 4–5),  
$\operatorname{median}a=5$ (IQR 5–6), and  
$\operatorname{median}\ell=6$ (IQR 5–6).
Thus, learners experience mildly positive affect, moderate arousal,
and above-neutral perceptions of learning benefit during most
interactions.

\paragraph*{Dominant discrete emotions (students only).}
The ten most frequent ensemble labels for students (Fig.~\ref{fig:top10_emotions})
cover 97.02\,\% of the corpus.  Three epistemic or low-polarity states
dominate: \textsc{neutral} appears in 3,910 turns (45.8\,\%), followed by
\textsc{confusion} in 1,891 (22.15\,\%) and \textsc{curiosity} in
1,351 (15.83\,\%).  Strongly negative emotions are comparatively rare:
\textsc{frustration} surfaces in 736 turns (8.62\,\%) and \textsc{anxiety}
in just 35 (0.41\,\%).

\vspace{0.5em}  % small breathing room before next subsection
% --------------------------------------------------------------
% --------------------------------------------------------------
\subsection{Temporal \& Transition Dynamics}
\label{sec:temporal}

\begin{figure}[t]
\centering
\includegraphics[width=0.82\linewidth]{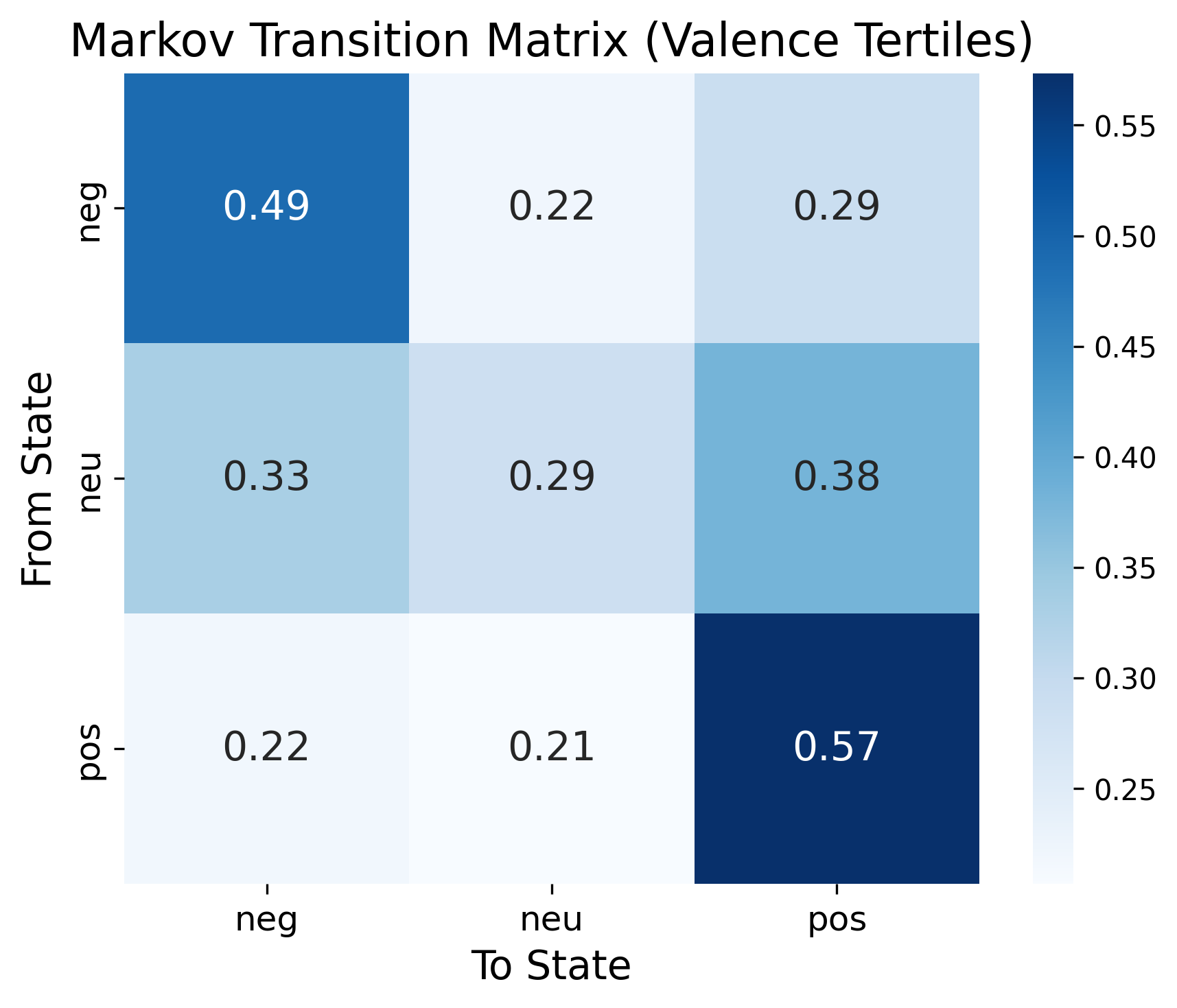}
\caption{Markov transition matrix ($3\times3$) for valence tertiles (student turns only; global tertiles; $\beta\!=\!1$ smoothing). Row $=$ current state, column $=$ next state.}
\label{fig:markov_matrix}
\end{figure}

\paragraph*{Markov state–transition structure.}
Figure~\ref{fig:markov_matrix} visualizes the first-order Markov transition matrix~$\mathbf{P}$ for \emph{student turns only}. Each turn is binned into negative, neutral, or positive valence using \emph{global} tertiles computed across all sessions.
Self-loop probabilities on the diagonal quantify \emph{emotional inertia}, the tendency for affect to persist once established. Off-diagonal entries capture cross-state shifts.

%%%%%%%%%%%%%%%%%%%%%%%% 
\section{Discussion}
\label{sec:discussion}
  Our results offer preliminary yet compelling insights into the affective contours of LLM-mediated tutoring. We interpret the findings through the lens of our research questions.

    \paragraph*{(RQ1) Ensemble-derived patterns.}
    The scalar valence–arousal–learning (VAL) distributions show that most student turns cluster around mildly positive valence ($\tilde{v}=5$), moderate arousal ($\tilde{a}=5$–6), and above-neutral learning appraisal ($\tilde{\ell}=6$). However, this central tendency belies substantial emotional diversity: confusion appears in 22.15\% of turns and frustration in 8.62\%, underscoring that affective friction persists even in scaffolded tutoring environments. These results nuance overly optimistic views of AI tutors and suggest that affect-regulation scaffolds may still be needed.

    \paragraph*{(RQ2) Affective transitions and persistence.} Our analysis reveals several notable patterns, outlined below.

\begin{enumerate}[label=\textbf{\arabic*.}, leftmargin=2.1em]
\item \textbf{Emotional inertia in AI–student interaction.} 
Within tutoring dialogues, learners are most likely to sustain a \emph{positive} state 
($P_{\text{pos}\to\text{pos}}=\mathbf{0.57}$), followed by \emph{negative} ($0.49$), 
with \emph{neutral} least stable ($0.29$). 
The corresponding expected dwell times $1/(1-P_{ss})$ are \textbf{2.33} (pos), 
\textbf{1.96} (neg), and \textbf{1.41} (neu) turns. 
This suggests that once a learner reaches a positive affective band in interaction with the AI tutor, 
they tend to remain there longer than in neutral or negative bands.

\item \textbf{Escape from negative affect is common and often direct.} Learners leave the negative band in \textbf{51\%} of turns ($1-0.49$). Direct rebounds to \emph{positive} are slightly more frequent than moves to \emph{neutral} ($P_{\text{neg}\to\text{pos}}=\mathbf{0.29}$ vs.\ $P_{\text{neg}\to\text{neu}}=0.22$), suggesting that improvements are often immediate rather than passing through neutrality.

\item \textbf{Outer-state asymmetry with a neutral tipping point.} Transitions between the extremes are not symmetric: $P_{\text{pos}\to\text{neg}}=0.22$ is lower than $P_{\text{neg}\to\text{pos}}=\mathbf{0.29}$, consistent with greater stability in the positive band. Meanwhile, \emph{neutral} behaves as a gateway with a slight positivity bias ($P_{\text{neu}\to\text{pos}}=0.38$ vs.\ $P_{\text{neu}\to\text{neg}}=0.33$) rather than a destination state ($P_{\text{neu}\to\text{neu}}=0.29$).
\end{enumerate}

% Analysis of within-session transitions shows that affect in AI–student dialogue exhibits short-horizon persistence. Positive states are the most stable (mean dwell $\approx$2.3 turns), negative somewhat less (2.0), and neutral brief (1.4). Importantly, recovery from negative affect is often rapid: over half of negative turns resolve immediately, with direct rebounds to positive ($0.29$) more frequent than detours through neutral ($0.22$). Neutral itself functions as a gateway rather than a destination, tipping upward more often than downward. Yet positive momentum is fragile—after a positive turn there remains a 43\% chance of reverting to non-positive on the next move. These dynamics imply that tutors should exploit neutral moments as launchpads, pair rapid repairs with consolidation, and scaffold after successes to sustain gains.

The following section synthesizes these findings and outlines concrete directions for future work.

\section{Conclusion}
\label{sec:conclusion}
This study presents one of the first large-scale analyses of affective dynamics in LLM-mediated tutoring. By leveraging ensemble affect annotations from three leading models: GPT-4o, Claude, and Gemini, we uncover how students’ emotional trajectories unfold over 16,986 conversational turns. 
Our results indicate that while most learners exhibit mildly positive valence and moderate arousal, epistemic emotions like confusion and curiosity remain prevalent. 
Our temporal analysis shows that students’ affective states in AI tutoring unfold in short bursts with rapid rebounds but fragile persistence, highlighting the need for tutor designs that provide timely scaffolds to repair negative affect and consolidate positive momentum.

The current analysis has several limitations that point to a clear agenda for future work. Most importantly, the absence of human-annotated gold data means we cannot yet determine how faithfully ensemble-derived labels reflect learners’ true affective states; a stratified, human-coded reference set will be essential for validation. In addition, while three frontier LLMs were used for annotation, we did not conduct ablation studies to quantify each model’s individual contribution or correctness—analyses that could guide more efficient and accurate ensemble design. Finally, our temporal modeling relied on a first-order Markov chain, which may miss longer-range dependencies and state-duration patterns; applying higher-order Markov models, hidden semi-Markov models, or neural sequence models could capture richer affective dynamics.

In closing, this work advances the conversation on how to responsibly integrate generative AI into education—not merely by improving test scores, but by attending to the emotional realities of learning.

\section*{Ethical Impact Statement} \label{sec:ethic}
This study is a secondary analysis of \textit{PyTutor} chat logs and was approved by the Institutional Review Boards of all three partner institutions (Protocol 2405001323 – Investigating Scaling a Personalized GenAI Tutor in Intro to Computing Courses). Students provided informed, opt-in consent and received no compensation. 

Ensemble affect scores are \emph{proxies}, not ground truth; we prohibit their use in high-stakes decisions without human review and will ship any deployed system with an opt-out toggle and visible affect-sensing indicator. 

All code, prompts, and analysis notebooks are available at \url{https://github.com/CharlieChenyuZhang/llm-ensemble-affective-tutoring} under an MIT licence, though raw logs cannot be shared due to FERPA constraints. Misclassifications may mislead, so our metrics should complement—not replace—human judgment.

%%%%%%%%%%%%%%%%%%%%%%%%%%%%%%%%%%%%%%%%%%%%%%%%%%%%%%%%%%%%%%%%
% ====== B I B L I O G R A P H Y ===============================
%%%%%%%%%%%%%%%%%%%%%%%%%%%%%%%%%%%%%%%%%%%%%%%%%%%%%%%%%%%%%%%%

\vspace{12pt}
\end{document}